\documentclass{article} % For LaTeX2e
\usepackage[accepted]{aistats2023}

\usepackage{listings}
\usepackage{microtype}
\usepackage{graphicx}
\usepackage{booktabs} % for professional tables

\usepackage[ruled,vlined]{algorithm2e}
\usepackage[utf8]{inputenc} % allow utf-8 input
\usepackage[T1]{fontenc}    % use 8-bit T1 fonts
\usepackage{url}            % simple URL typesetting
\usepackage{booktabs}       % professional-quality tables
\usepackage{amsfonts}       % blackboard math symbols
\usepackage{nicefrac}       % compact symbols for 1/2, etc.
\usepackage{microtype}      % microtypography
\usepackage{amsmath}
\usepackage{amssymb}
\usepackage{theorem}
\usepackage{algpseudocode}
\usepackage{graphicx}
\usepackage{stackengine}
\usepackage{textcomp}
\usepackage{multirow}
\usepackage{footnote}
\usepackage{subcaption}
\usepackage{xcolor}
\usepackage[pagebackref=true]{hyperref} 
\hypersetup{colorlinks,allcolors=black}
\renewcommand*\backref[1]{\ifx#1\relax \else (Cited on #1) \fi}

\hypersetup{colorlinks=true,
            citecolor=[rgb]{0.69,0.16,0.16},
            linkcolor=[rgb]{0.69,0.16,0.16},
            urlcolor=[rgb]{1,0,1}}

\emergencystretch=\maxdimen
\renewcommand{\leq}{\ensuremath{\leqslant}}

\renewcommand{\ge}{\ensuremath{\geqslant}}
\renewcommand{\le}{\ensuremath{\leqslant}}
\newcommand{\menge}[2]{\big\{{#1}~\big |~{#2}\big\}} 
\newcommand{\scal}[2]{{\left\langle{{#1}\mid{#2}}\right\rangle}}

\newcommand{\NN}{\ensuremath{\mathbb N}}
\newcommand{\RR}{\ensuremath{\mathbb R}}
\newcommand{\RP}{\ensuremath{\left[0,+\infty\right[}}

\newcommand{\RPP}{\ensuremath{\,\left]0,+\infty\right[}}

\newcommand{\prox}{\ensuremath{\mathrm{relax}}}
\newcommand{\proj}{\ensuremath{\mathrm{proj}}}

\newcommand{\pinf}{\ensuremath{+\infty}}

\newcommand{\minimize}[2]{\ensuremath{\underset{\substack{{#1}}}%
{\text{\rm minimize}}\;\;#2 }}

\newcommand{\argmind}[2]{\ensuremath{\underset{\substack{{#1}}}%
{\text{\rm argmin}}\;\;#2 }}

\newtheorem{theorem}{Theorem}%[section]

\newtheorem{proposition}[theorem]{Proposition}
\theoremstyle{plain}{\theorembodyfont{\rmfamily}%
}
\theoremstyle{plain}{\theorembodyfont{\rmfamily}%
\newtheorem{definition}[theorem]{Definition}}
\theoremstyle{plain}{\theorembodyfont{\rmfamily}%
}
\theoremstyle{plain}{\theorembodyfont{\rmfamily}%
\newtheorem{problem}[theorem]{Problem}}
\theoremstyle{plain}{\theorembodyfont{\rmfamily}%
}

\renewcommand{\eqref}[1]{Eq.~(\ref{#1})}

\definecolor{codegreen}{rgb}{0,0.6,0}
\definecolor{codegray}{rgb}{0.5,0.5,0.5}
\definecolor{codepurple}{rgb}{0.58,0,0.82}
\definecolor{backcolour}{rgb}{0.95,0.95,0.92}

\lstdefinestyle{mystyle}{
    backgroundcolor=\color{backcolour},   
    commentstyle=\color{codegreen},
    keywordstyle=\color{magenta},
    numberstyle=\tiny\color{codegray},
    stringstyle=\color{codepurple},
    basicstyle=\ttfamily\footnotesize,
    breakatwhitespace=false,         
    breaklines=true,                 
    captionpos=b,                    
    keepspaces=true,                 
    numbers=left,                    
    numbersep=5pt,                  
    showspaces=false,                
    showstringspaces=false,
    showtabs=false,                  
    tabsize=2
}

\begin{document}

\twocolumn[

\aistatstitle{CertViT: Certified Robustness of Pre-Trained Vision Transformers}

\aistatsauthor{Kavya Gupta \And Sagar Verma}

\aistatsaddress{ Universit\'{e} Paris-Saclay, CentraleSup\'{e}lec\\ Inria, Centre de Vision Num\'{e}rique \And Universit\'{e} Paris-Saclay, CentraleSup\'{e}lec\\ Inria, Centre de Vision Num\'{e}rique \\ Granular AI  } ]

% The \author macro works with any number of authors. There are two commands
% used to separate the names and addresses of multiple authors: \And and \AND.
%
% Using \And between authors leaves it to \LaTeX{} to determine where to break
% the lines. Using \AND forces a linebreak at that point. So, if \LaTeX{}
% puts 3 of 4 authors names on the first line, and the last on the second
% line, try using \AND instead of \And before the third author name.

\newcommand{\fix}{\marginpar{FIX}}
\newcommand{\new}{\marginpar{NEW}}

\begin{abstract}

Lipschitz bounded neural networks are certifiably robust and have a good trade-off between clean and certified accuracy. Existing Lipschitz bounding methods train from scratch and are limited to moderately sized networks (< 6M parameters). They require a fair amount of hyper-parameter tuning and are computationally prohibitive for large networks like Vision Transformers (5M to 660M parameters). Obtaining certified robustness of transformers is not feasible due to the non-scalability and inflexibility of the current methods. This work presents CertViT, a two-step proximal-projection method to achieve certified robustness from pre-trained weights. The proximal step tries to lower the Lipschitz bound and the projection step tries to maintain the clean accuracy of pre-trained weights. We show that CertViT networks have better certified accuracy than state-of-the-art Lipschitz trained networks. We apply CertViT on several variants of pre-trained vision transformers and show adversarial robustness using standard attacks. Code : \url{https://github.com/sagarverma/transformer-lipschitz}
\end{abstract}

\section{Introduction}

\begin{figure}[!ht]
     \centering
     \begin{subfigure}[b]{0.22\textwidth}
         \centering
         \includegraphics[width=0.9\textwidth]{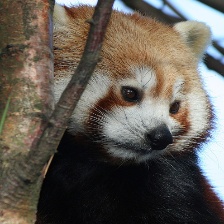}
         \caption{Input Image}
         \label{fig:panda}
     \end{subfigure} 
     \quad
     \begin{subfigure}[b]{0.22\textwidth}
         \centering
         \includegraphics[width=0.9\textwidth]{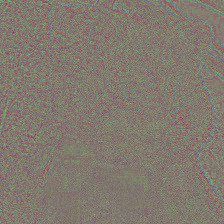}
         \caption{Perturbation for ViT}
         \label{fig:panda_vit}
     \end{subfigure} \\
     \begin{subfigure}[b]{0.22\textwidth}
         \centering
         \includegraphics[width=0.9\textwidth]{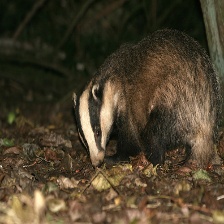}
         \caption{ViT Prediction}
         \label{fig:panda_false}
     \end{subfigure}
     \quad
     \begin{subfigure}[b]{0.22\textwidth}
         \centering
         \includegraphics[width=0.9\textwidth]{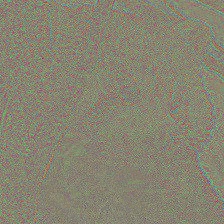}
         \caption{Perturbation for CertViT}
         \label{fig:panda_certvit}
     \end{subfigure}
        \caption{\textbf{Transformers vs Certified Transformers}. Shown here are adversarial perturbations ($\ell_2$) computed for a transformer and its certified version using the proposed method (CertViT). While , ViT fails under perturbation and predicts the input image of \textit\textbf{panda} as \textit\textbf{badger}, CertViT predicts it correctly.}
        \label{fig:panda}
\end{figure}

Deep neural networks (DNNs) are vulnerable to adversarial examples \cite{szegedy2013intriguing}, where the perturbations to the input are constructed deliberately to confuse the classifier. This erratic behavior of DNNs is possible even when the model accuracy on the clean samples is excellent. Numerous heuristic solutions for adversarial defenses have been proposed in the literature, but these solutions are often broken using more carefully crafted stronger attacks \cite{athalye2018obfuscated}. This suggests that these methods provide only empirical robustness without any formal guarantees. Formal guarantees are of much importance in mission- and safety-critical applications. For such scenarios, we need certified defenses with formal robustness guarantees that any norm-bounded adversary will not be able to alter the network predictions. One such line of work providing certified robustness concentrates on estimating and subsequently bounding the Lipschitz constant of a neural network. This can be done in both global and local contexts. Global Lipschitz constant is computationally cheap and scalable but often loose and hence tends to over-regularize the training and reduce the network capacity to learn. In comparison, local Lipschitz estimates are tight since it uses information in the local neighborhood of the input data but are hard to estimate and time-consuming. State-of-the-art methods obtain global and local Lipschitz bounded networks when training from scratch. These methods have been applied to moderately sized networks comprising convolutional and dense layers. The largest network experimented in \cite{virmaux2018lipschitz} consists of 6M parameters. These methods become untenable for deeper and more complex architectures such as transformers.

Transformers proposed by \cite{vaswani2017attention} for machine translation has become the state-of-the-art method in many NLP tasks. BERT \cite{devlin2018bert} uses a denoising self-supervised pre-training task, and Generative Pre-trained Transformer (GPT) \cite{radford2018improving} uses language modeling as its pre-training task. In \cite{dosovitskiy2020image}, the authors have proposed Vision Transformer (ViT) with very few changes to the original transformer. They proposed to split the input image into small patches and provide the sequence of linear embedding of the patches as input. Data-efficient ViT (DeiT) reduces the dependence of ViTs on large-size datasets. DeiT improved the efficiency of ViTs by employing data augmentations and regularization. Swin Transformer presented a hierarchical transformer that reduces the number of tokens through patch merging while using attention in non-overlapping local windows. Many different Vision Transformer architectures are proposed for various tasks, but the attention layer remains the backbone of most of them. It has to be noted that, for ViT, the number of parameters tends to be very large, so it is computationally expensive to train them from scratch. Instead, researchers use their pre-trained models for fine-tuning desired tasks or datasets.

Although there have been many works around the empirical robustness of transformers and several adversarial attacks specifically designed for fooling transformers, the internal properties relating to robustness are not well studied. As mentioned earlier, one way of providing theoretical guarantees of robustness is through Lipschitz bounds. In \cite{kim2021lipschitz}, authors have proved that the standard dot product (DP) self-attention is not Lipschitz continuous. They provide a Lipschitz continuous $\ell_2$ distance based self-attention as the replacement of DP. They show the efficacy of their transformer-based architecture on language modeling tasks in terms of performance. To the best of our knowledge, the certified robustness of pre-trained transformers has not been investigated. Transformers being large networks efficacy of current certifiable robust training methods is unknown. This paper focuses on obtaining the certified robustness of large pre-trained transformers. Our main contributions are summarized as follows:

\begin{itemize}
    \item We present a method to constrain the Lipschitz bounds of pre-trained transformers using the Douglas-Rachford method. The algorithm consists of two steps: lowering the Lipschitz constant of the network and maintaining good clean accuracy.
    
    \item We first compare the effectiveness of our proposed method \emph{CertViT} in certifying standard convolution and linear layer-based models on different datasets and compare it to various Lipschitz bound-based certifying methods. 
    
    \item For transformers, we adopt a Lipschitz continuous L2 self-attention from pre-trained DP attention networks, making them Lipschitz continuous. We then demonstrate the effectiveness of our method in certifying different transformer architectures.

    \item We demonstrate the computational cost aspects of training a robust transformer from scratch using existing methods and compare them with our proposed method.
    
    \item We report results on several varieties of ImageNet-1K pre-trained vision transformers using our proposed method. We also discuss the limitations associated with it.
\end{itemize}

% {\color{blue} The rest of the paper is organized as follows. In the next section, we discuss literature related to various aspects of certified robustness. Section~\ref{sec:proposed} discusses in detail our proposed approach. Section~\ref{sec:experiments} provides our experimental setup with results comparison with state-of-the-art. We also discuss insights on the certification of transformers. Section~\ref{sec:conclusion} concludes and summarizes our work.}

\section{Related Work} \label{sec:soa}

In \cite{virmaux2018lipschitz}, the problem of computing the exact Lipschitz constant of a differentiable function is shown to be NP-hard; hence most works focus on finding the tightest upper bound. Virmaux et al. \cite{virmaux2018lipschitz} proposed the first generic algorithm (AutoLip) for upper bounding the Lipschitz constant of any differentiable function. RecurJac \cite{zhang2019recurjac} is a recursive algorithm analyzing the local Lipschitz constant in a neural network using a bound propagation \cite{zhang2019towards}. FastLip \cite{weng2018towards} is a weaker form of RecurJac. SDP-based Lipschitz Constant estimates were explored in \cite{fazlyab2019efficient}. 
Using the non-expansiveness property of the activation function, the optimization problem of estimating a Lipschitz constant is recast as solving a semi-definite positive programming problem, and this estimate is limited to $\ell_2$ perturbations. Jordan et al. \cite{jordan2020exactly} compute tight Lipschitz bound using mixed integer programming. In \cite{latorre2020lipschitz}, authors proposed a polynomial constrained optimization (LipOpt) based technique for estimating the Lipschitz constant. This estimate can be used for any $\ell_p$ norm for input and output perturbations, but it is valid for a neural network with a single output. These estimates though tight are only limited to small and pre-trained models since it is difficult to parallelize the optimization solver and make it differentiable for training. 

Utilizing the Lipschitz constant to certify robustness has been studied in several instances in the literature. Neural networks trained without any robustness constraint usually have very large global Lipschitz constant bounds \cite{szegedy2013intriguing}, most existing works train the network with the criterion that promotes small Lipschitz bound. Cisse et al.\cite{cisse2017parseval} designed networks with orthogonal weights whose Lipschitz constants are exactly 1. Miyato et al.\cite{miyato2018spectral} showed control on the Lipschitz constant using spectral normalization for GANs and in \cite{serrurier2021achieving}, authors learn 1-Lipschitz networks using hinge regularisation loss for binary classifiers. %There has also been prior work seeking to use Lipschitz bounds, either global or local, during training to promote robustness. Global Lipschitz bounds are often easy to compute but they tend to over-regularize the network training and decrease its clean accuracy. 
Lipschitz Margin training (LMT) \cite{tsuzuku2018lipschitz} trains networks that are certifiably robust by constructing a loss on worst logits which are calculated using the global Lipschitz bounds. It adds $\sqrt{2}\epsilon L_{glob}$, where $\epsilon$ is the perturbation radius to be certified, to all logits other than that corresponding to the ground-truth class. Box constrained propagation (BCP) \cite{lee2020lipschitz} achieves a tighter outer bound than the global Lipschitz-based outer bound, by considering local information via interval bound (box) propagation. They also compute the worst-case logit based on the intersection of a (global) ball and a (local) box. GloRo \cite{leino2021globally} bounds the upper bounds on the worst margins using the global Lipschitz constant. It constructs a new logit with a newly constructed class called the bottom class and determines if the sample can be certified. 
Local-Lip \cite{huang2021training} utilizes the interactions between activation functions (e.g. ReLU, MaxMin) and weight matrices. By eliminating the corresponding rows and columns where the activation function output is constant they guarantee a lower provable local Lipschitz bound than the global Lipschitz bound for the neural network. Gouk et al. \cite{gouk2021regularisation} proposed a relaxed Lipschitz constrained training method in which neural network optimizers can be adapted with a projection step to lower the Lipschitz constant. This is achieved by normalizing the parameters of a neural network by its spectral norm at every training step.

\textbf{On robustness of Transformers} - Adversarial robustness of Vision Transformers (ViT) has been studied in \cite{bhojanapalli2021understanding,mahmood2021robustness,benz2021adversarial} and they compare different ViTs to their ResNet counterparts. \cite{mahmood2021robustness} studied commonly used adversarial attacks and concluded attacks do not transfer readily between transformers and ResNets. They also proposed a new attack SAGA(Self-Attention Gradient attack) for attacking an ensemble of transformers and CNNs. Benz et al. \cite{benz2021adversarial} did a frequency analysis which suggests that the most robust ViT architectures tend to rely more on low-frequency features compared to CNNs. In \cite{bhojanapalli2021understanding} experiments suggest that ViTs are more robust to the removal of any single layer. An interesting observation from the paper suggests a high correlation in the later layers of ViTs, indicating a larger amount of redundancy. 

\textbf{Adversarial robustness of pre-trained classifiers} - Salman et al.\cite{salman2020denoised} proposed a method called denoised smoothing to make pre-trained classifiers robust without any retraining or fine-tuning. It prepends a custom-trained Gaussian denoiser to the pre-trained classifier and applies randomized smoothing \cite{cohen2019certified} to the whole network resulting in a certifiably robust classifier. It applies to $\ell_2$ bounded perturbations and is not easily extensible to other perturbation models. \cite{norouzzadeh2021empirical} offers a higher empirical robust accuracy than denoised smoothing, eliminating the need for multiple queries per sample and reducing the high computational cost of multiple forward passes during inference time. \cite{norouzzadeh2021empirical} show the applicability of their approach to $\ell_{\infty}$ perturbations as well. Such methods also provide a high degree of empirical robustness but do not provide any formal guarantees of robustness.

\section{Proposed Method}\label{sec:proposed}
\vspace{-1em}
\subsection{Lipschitz Continuity}

\begin{definition}
    \textit{Lipschitz constant of a function $f$ is an upper bound on the ratio between the output and the input variations of a function $f$.
    If $L\in [0,+\infty[$ is such that,
    for every input $x\in \RR^{N_0}$ and perturbation $z\in \RR^{N_0}$,
    \begin{equation}
    \|f(x+z)-f(x)\|\leq L\|z\|
    \end{equation}
    then  $L$ is a Lipschitz constant of $f$.
     $\|.\| $ denotes the standard Euclidean norm, but any other norms can be applied.
     If $X$ is defined as the $\epsilon$-ball at point $x$, i.e., $X =\{{x^{'}|\|x - x^{'}\| \leq \epsilon\}}$, then $L$ is the local Lipschitz constant of $f$ at $x$.}
\end{definition}

The first upper bound on the Lipschitz constant of a neural network was derived by analyzing the effect of each layer independently and considering a product of the resulting spectral norms \cite{goodfellow2014explaining}. 
For an $m$-layered neural network the global Lipschitz Bound is given as follows: 

 \begin{equation}\label{e:trivupbound}
     \overline{L}_{glob}
     \leq
 \|W_m\|_{\rm S} \|W_{m-1}\|_{\rm S}\cdots \|W_{1}\|_{\rm S}.
 \end{equation}
\begin{align*}
    (\forall i \in \{1,\ldots,m\})\quad 
    x_{i} = \phi_i(W_i x_{i-1}+b_i),
\end{align*}

where, at the $i^{\rm th}$ layer, $x_{i-1}\in \RR^{N_{i-1}}$ designates
the input, $x_i\in \RR^{N_i}$ the output,
$W_i\in \RR^{N_i\times N_{i-1}}$ is
the weight matrix, $b_i\in \RR^{N_i}$ is the bias vector, and $\phi_i\colon \RR^{N_i}\to \RR^{N_i}$ is the activation operator, and $\|\cdot\|_{\rm S}$ denotes the spectral matrix norm and is equal to the maximum singular value of $W$ and note bias don't contribute to the Lipschitz bound. An important assumption is that the activation operators are nonexpansive, i.e., 1-Lipschitz. This assumption is satisfied for all the standard choices of activation operators ReLU, tanh etc.%Although easy to compute this bound is over-pessimistic since it ignores the non-linearities of the deep neural network. 

\subsection{Linear layers and variational principles} \label{s:varprinc}
A single layer linear network without any bias can be defined as:
\begin{equation}\label{e:onelabel}
y = R(Wx)
\end{equation}
where $x\in \RR^{M}$ is the input, 
$y\in \RR^N$ the output, $W\in \RR^{N\times M}$
is the weight matrix, and $R$ is a nonlinear activation operator from $\RR^N$ to $\RR^N$. Combettes et al. \cite{Combettes2020varineq,combettes2020lipz} recently showed that most neural network activation functions are proximity operators of convex functions. This shows that there exists a proper lower-semicontinuous convex function $f$ from $\RR^N$ to $\RR\cup \{+\infty\}$ such that $R=\prox_f$. We recall that $f$ is a proper lower-semicontinuous convex function if the area overs its graph,
its epigraph
$\menge{(y,\xi)\in \RR^N\times \RR}{ f(y) \le \xi}$, 
is a nonempty closed convex set. For such a function
the proximity operator of $f$ at $z\in \RR
^N$ \cite{moreau1962} is the unique point defined as 
\begin{equation}
  \prox_f(z) = 
  \argmind{p\in \RR^N}{\frac12 \|z-p\|^2+f(p)}.
\end{equation}
It follows from standard subdifferential calculus that 
\eqref{e:onelabel} can be re-expressed as the following 
inclusion relation:
\begin{equation}\label{e:varincl}
Wx-y \in \partial f(y),
\end{equation}
where $\partial f(y)$ is the Moreau subdifferential of
$f$ at $y$ defined as
\begin{equation}
    \partial f(y) = \menge{t\in \RR^N}{(\forall z \in \RR^N) f(z) \ge f(y)+\scal{t}{z-y}}.
\end{equation}
The subdifferential constitutes a useful extension of the notion of differential, which is applicable to nonsmooth functions.
The set $\partial f(y)$ is closed and convex and, if $y$ satisfies \eqref{e:onelabel}, it is nonempty.
The distance to this set of a point $z\in \RR^N$ 
is given by
\begin{equation}
    d_{\partial f(y)}(z) = \inf_{t\in \partial f(y)}
    \|z-t\|.
\end{equation}
We thus see that the subdifferential inclusion in \eqref{e:varincl}
is also equivalent to
\begin{equation}
d_{\partial f(y)}(Wx-y) = 0.
\end{equation}
Therefore, a suitable accuracy measure for approximated values of the layer parameters $(W)$ is $d_{\partial f(y)}(Wx-y)$.

\subsection{Constraining Lipschitz of a layer}

Aiming for a tight Lipschitz bound of a network consists of making the Lipschitz constant of individual layers of the network small independently while keeping an overall satisfactory accuracy. We assume that, for a given layer, a training sequence of input/output pairs is available which results from a forward pass performed on the original/pre-trained network for some input dataset of length $K$. The training sequence is split into $J$ mini-batches of size $T$ so that $K = J T$. The $j$-th mini-batch with  $j\in \{1,\ldots,J\}$ is denoted by $(x_{j,t},y_{j,t})_{1\le t\le T}$. To lower the Lipschitz constant of the network, we propose to solve the following constrained optimization problem

\begin{problem}\label{p:opt}
%We want to
\begin{equation}
\minimize{W\in C}{L(W)}
\end{equation}
with

\begin{align}
C = \Big\{&W\in \RR^{N\times M} \mid
(\forall j\in \{1,\ldots,J\})\nonumber  \\
& \qquad \sum_{t=1}^T d_{\partial f(y_{j,t})}^2(Wx_{j,t}-y_{j,t}) \le T\eta\Big\},
\label{e:constac}
\end{align}

where $L$ is a Lipschitz measure defined on $\RR^{N\times M}$, $\eta \in \RP$ is defined accuracy tolerance, and $d_{\partial f(y)}(Wx-y)$ is a suitable accuracy measure for approximated values of the layer parameters $W$.
\end{problem}

Since, for every $j\in \{1,\ldots,J\}$, the function
$W\mapsto \sum_{t=1}^T d_{\partial f(y_{j,t})}^2(Wx_{j,t}-y_{j,t})$ is continuous and convex, $C$ is a closed and convex subset of $\RR^{N\times M}$. In addition, this set is nonempty when there exist $\overline{W}\in \RR^{N\times M}$ such that, for every $j\in \{1,\ldots,J\}$ and $t\in \{1,\ldots,T\}$,
\begin{equation}
d_{\partial f(y_{j,t})}^2(\overline{W}x_{j,t}-y_{j,t}) = 0.
\end{equation}
This condition is satisfied when $\overline{W}$ are the parameters of the pre-trained layer. A standard choice for such a function is the $\ell_1$-norm of the matrix elements, $L = \|\cdot\|_1$. 

A standard proximal method for solving Problem \ref{p:opt} is the Douglas-Rachford algorithm \cite{pll1979splitting,plc2007drsplit}.This algorithm alternates between a proximal step ({$\prox_\beta$}) aiming at lowering the Lipschitz value of the weight matrix and a projection step ({$\proj_C$}) to maintain original accuracy of the pre-trained network. This algorithm can be written as :

    \begin{algorithm}[ht]
%        \SetKwInOut{Require}{Require}
        \SetKwInOut{Initialize}{Initialize}
        \caption{ CertViT: Constraining a Layer}
        \label{a:DR}
        %\Require{}
        \Initialize{$\widehat{W}_0\in \RR^{N\times M}$}
        \For{$n=0,1,\ldots$}{
            $W_n= \prox_\beta(\widehat{W}_n)$ \\
            $\widetilde{W}_n=\proj_C(2W_n-\widehat{W}_n, \eta)$\\
            $\widehat{W}_{n+1} = \widehat{W}_n + \lambda_n (W_n-\widetilde{W}_n)$\\
        }
    \end{algorithm}
    
The Douglas-Rachford algorithm uses positive parameters $\beta$ and $(\lambda_n)_{n\in\NN}$. Throughout this article, $\proj_S$ denotes the projection onto a nonempty closed convex set $S$. 

\begin{proposition}\cite{plc2007drsplit}
Assume that Problem \ref{p:opt} has a solution and that there exists $\overline{W}\in C$
such  $\overline{W}$ is a point in the interior of the domain of $h$.
Assume that $\beta \in \RPP$ and $(\lambda_n)_{n\in\NN}$ in $]0,2[$ is
such that $\sum_{n\in \NN} \lambda_n (2-\lambda_n) = \pinf$.
Then the sequence
$W_{n\in \NN}$ generated by Algorithm~\ref{a:DR} converges to
a solution to Problem~\ref{p:opt}.
\end{proposition}

The projection step $\prox_\beta$ reduces the magnitude of each parameter or element of the input matrix $W_n$ by $\beta$ (i.e $(\text{sign}(W_n) * (\text{abs}(W_n) - \beta))$). This is equivalent to making a matrix sparse also known as magnitude pruning \cite{verma2021sparsifying}. In turn, since the convex set $C$ has an intricate form, an explicit expression of $\proj_C$ does not exist. Finding an efficient method for computing this projection for large datasets thus constitutes the main challenge in the use of the above Douglas-Rachford strategy, which we will discuss in the next section.

\subsection{Maintaining clean accuracy}

For every mini-batch index $j\in \{1,\ldots,J\}$, we define the following convex function:
\begin{multline}\label{e:defcj}
(\forall W\in \RR^{N\times M}) \\
    c_j(W) = \sum_{t=1}^T d_{\partial f(y_{j,t})}^2(Wx_{j,t}-y_{j,t})-T\eta.
\end{multline}
Note that, for every $t \in \{1,\ldots,T\}$ and $j\in \{1,\ldots,J\}$, function $c_j$ is differentiable and its gradient at  $W\in \RR^{N\times M}$ is given by
\begin{equation}
    \nabla c_j(W) = 2\sum_{t=1}^T  (Wx_{j,t}-y_{j,t}) x_{j,t}^\top.
    \label{e:gradc}
\end{equation}

Weight parameters belong to $C$ if and only if it lies in the intersection of the 0-lower level sets of the functions $(c_j)_{1\le j \le J}$.
To compute the projection of some $W\in \RR^{N\times M}$ onto this intersection, we use 
Algorithm \ref{a:proj} 
($\|\cdot\|_{\rm F}$ denotes here the Frobenius norm).

\begin{algorithm}[!ht]
%        \SetKwInOut{Require}{Require}
        \SetKwInOut{Initialize}{Initialize}
        \caption{Mini-batch algorithm for computing $\proj_C(W, \eta)$ }
        \label{a:proj}
        
        \Initialize{$W_0=W$}
        \For{$n=0,1,\ldots$}{
            Select a batch of index $j\in \{1,\ldots,J\}$\\
            \If{$c_{j}(W) > 0$}
            {
            $\delta W = \frac{c_{j}(W)\, \nabla c_{j}(W)}{\|\nabla c_{j}(W)\|_{\rm F}^2}$\\
            $\pi_n = ((W_0-W)^\top \delta W)$\\
            $\mu_n = \|W_0-W\|_{\rm F}^2$\\
            $\nu_n = \|\delta W\|_{\rm F}^2$\\
            $\zeta_n=\mu_n \nu_n-\pi_n^2$\\
            \If{$\zeta_n=0$ and $\pi_n \ge 0$}{
            $W = W + \delta W$\\
            }
            \ElseIf{$\zeta_n > 0$ and $\pi_n \nu_n\ge \zeta_n$}
            {
            $W = W_0 + (1+\frac{\pi_n}{\nu_n}) \delta W$\\
            }
            \Else{
            $W = W + \frac{\nu_n}{\zeta_n} (\pi_n(W_0-W)-\mu_n \delta W)$\\
            }
            }
            \Else{
            $W_{n+1} = W_n$\\
            }
        }
    \end{algorithm}

This algorithm has the advantage of proceeding in a mini-batch manner. The simplest rule is to use each mini-batch once within $J$ successive iterations of the algorithm so that they correspond to an epoch. The proposed algorithm belongs to the family of block-iterative outer approximation schemes for solving constrained quadratic problems \cite{plc2003block}. 
% The convergence of the sequence $W_{n\in \NN}$ generated by Algorithm \ref{a:proj} to $\proj_C(W)$ is thus guaranteed. 
One of the main features of this algorithm is that it does not require performing any projection onto the 0-lower level sets of the functions $c_j$, which would be intractable due to their expressions. Instead, these projections are implicitly replaced by subgradient projections, which are much easier to compute in our context. 

% \textcolor{blue}{Do we need to write regarding the soft-thresholding things? about fine-tuning? that we freeze the zero. if would be good to summarize the training procedure a little bit stressing on "good" pretrained weights. }

%subsection{Dealing with Attention Layer}

\subsection{CertViT for a Network}

\begin{algorithm}
\caption{Parallel CertViT for multi-layer network}
\label{a:parallel}
\SetKwInput{KwInput}{Input}
\SetKwInput{KwOutput}{Output}
\KwInput{input sequence $X\in \RR^{M\times K}$, $\beta$ magnitude for soft-thresholding, error tolerance parameter $\eta > 0$, $\lambda$ is update rate, weight matrices $W^{1},\dots,W^{L}$}  

$Y^{0} \leftarrow X$ \\
\For{$l=1,\dots,L$}{
    $Y^{l} = R_l(W^{l} Y^{l-1})$ \\ 
    $\widehat{W}^{l} \leftarrow \text{CertViT}(\beta, \eta, \lambda, W^{l},Y^{l},Y^{l-1})$
}
 $\widehat{W}^{1},\dots,\widehat{W}^{L} \leftarrow \text{fine-tune}(\widehat{W}^{1},\dots,\widehat{W}^{L}, X, Y)$ \\
\KwOutput{$\widehat{W}^{1},\dots,\widehat{W}^{L}$}
\end{algorithm}

Algorithm \ref{a:parallel} describes how we make use of CertViT for a transformer. We use a pre-trained transformer and the training sequence to extract layer-wise input-output features. Then we apply CertViT on individual layer $l$ by passing parameter $\beta$, $\eta$, and $\lambda$ for magnitude based soft-thresholding, error tolerance, and Douglas-Rachford update rate, respectively. $R_l$ is the activation function just after the layer $l$. Layer parameters $W^{l}$ and input-output features $(Y^{l-1},Y^{l})$ are extracted and passed to Algorithm~\ref{a:DR}. The benefit of applying CertViT to each layer independently is that we can run CertViT on all the layers of a network in parallel. This reduces the time required to process the whole network and compute resources are optimally utilized. Once all layers have been constrained using CertViT, we finally do a fine-tuning on training set to overcome any accuracy lost during the process. This is similar to retraining after magnitude based pruning. 

\subsection{Computing Lipschitz constant of Transformers}

%\subsubsection{Multi-Headed Self-Attention}
Transformer model comprises of fully connected, convolutional, and self-attention layers. Though the estimates of the Lipschitz constant of fully connected and convolutional layers are now widely studied in the literature, self-attention layers are still under-explored in terms of Lipschitz bounds and hence their certified robustness is of question. In \cite{kim2021lipschitz} authors %went beyond fully connected and convolutional layers to
investigate the Lipschitz bounds of self-attention layer.% which is the backbone of the transformer model. 
They prove the standard DP self-attention used in all the current transformer models is not Lipschitz. They propose an alternative $L2$ Multi headed self-attention ($L2$-MHA).% $X \in \RR^{N \times D}$ is the concatenation of $N$ patches of $D$ dimension and is input to the self-attention layer. Then L2-MHA map: $F:\RR^{N\times D} \rightarrow \RR^{N\times D}$, where $N$ is the input sequence length, $H$ is the number of heads, $D$ is the embedding dimension. $W^{K,h},W^{V,h},W^{Q,h}$  are the key, value and query weight matrices respectively:\\
% \begin{align*}
%      F(X)= \Big[ S_1(X)W^{V,1},\dots, S_H(X)W^{V,H} \Big]W^O 
% \end{align*}
%  where $S_h(X) = P^hXA_h$ , $W^{V,h} \in \RR^{D \times D/H}$ , $W^{Q,h},W^{K,h}  \in \RR^{D \times D/H}$, $W^{O} \in \RR^{D \times D}$ , $A_h = W^{Q,h}W^{Q,h^{T}}/\sqrt{D/H} \in \RR^{D \times D} $
%  $P_h$ for each head is defined as 
% \begin{equation}
%     P^h = softmax\Bigg(-\frac{\|XW^{Q,h}\|_{row}^2\mathbbm{1}^T-2XW^{Q,h}(XW^{K,h})^T + \mathbbm{1}\|XW^{K,h}\|_{row}^{2^{T}}}{\sqrt{D/H}}\Bigg)
% \end{equation}

% The $L2$ and $L_\infty$ Lipschitz bounds  are given as:
%  \begin{equation}
%  L_{\infty}(f) \leq \Bigg(4\phi^{-1}(N-1) + \frac{1}{\sqrt{D/H}}\Bigg) \|W^{O^T}\|_\infty \max\limits_h \|W^{Q,h}\|_\infty \max\limits_h\|W^{Q,h^T}\|_\infty \max\limits_h\|W^{V,h^T}\|_\infty
%  \end{equation}
 
%  \begin{equation}
%  L_{2}(f) \leq \frac{\sqrt{N}}{\sqrt{D/H}}(4\phi^{-1}(N-1))\Bigg(\sum\limits_h\sqrt{\|W^{Q,h}\|_2^2\|W^{V,h}\|_2^2}\Bigg)\|W^{O}\|_2
%  \end{equation}
%  where $\phi(x) := xe^{(x+1)}$  is an invertible univariate function on $x > 0$. 
These estimates are Lipschitz continuous for the condition that query and key weight matrices are shared for each head in multi-headed self attention module. %More details can be found in \cite{kim2021lipschitz}.
The estimates presented imply transformers have intrinsically high Lipschitz constant due to the self-attention module and naively restricting them to attain lower Lipschitz constant makes them difficult to learn leaving both clean and certified accuracy very low. Since this form of self-attention is Lipschitz continuous we adapt it to learn the ViTs from scratch and also adapt the DP-attention to $L_2$ attention weights in pre-trained transformers.

%\subsubsection{Patch Embedding and MLP layers}
The patch embedding layer is used for generating patches and translating the patches to fixed dimension with their positional embeddings. This layer can be constructed using convolutional or dense layers. The output of the patch embedding layer serves as the input to the transformer encoder. The output of this encoder is fed to the MLP layer. The Lipschitz bounds for the weights of both of these types of layers can be estimated using spectral norm via the power iteration method. The MLP layer consists of two dense layers. The first dense layer of MLP is followed by the GeLU activation function. GeLU is not 1-Lipschitz, but Lipschitz continuous with value 1.12 which can be trivially obtained (Appendix ~\ref{gelu}). So we calculate the Lipschitz bound for the MLP layer in the transformer as $1.12\|W\|_s$.
%\subsubsection{MLP}

%\textcolor{red}{Layer Normalisation}

% \subsection{Lipschitz Training}

%  We use the same training strategy to create some baselines to compare with our proposed method. We denote this training strategy as $\mathcal{L_{\gamma}}$, where $\gamma$ is the relaxation parameter applied to relax the spectral norm division effect on weight to make the network have more learning capacity. 

\begin{table*}[!ht]
    \centering
    \begin{tabular}{l c c c c c c}
    \toprule
         \textbf{Method (Params)} & \textbf{Model} & \textbf{Clean (\%)} & \textbf{PGD (\%)} & \textbf{Cert. (\%)} & \textbf{Lip.} & \textbf{FLOPs ($\times10^{13}$)} \\
         \midrule
         \multicolumn{7}{c}{\textbf{MNIST} ($\epsilon = 1.58$)} \\
         \midrule
         \textbf{Standard (1973536)} & 4C3F & \textbf{99.3} & 45.9 & 0.0 & $2.5\times10^3$ & 6.1 \\
         \textbf{BCP} & 4C3F & 92.4 & 65.8 & 44.9 & 6.9 & 18.2 \\
         \textbf{GloRo} & 4C3F & 97.0 & 68.9 & 50.1 & 2.3 & 30.3 \\
         \textbf{Local-Lip} & 4C3F & 96.2 & 78.2 & 55.8 & \textbf{0.7} & 17.6 \\
         \textbf{CertViT (Ours)} & 4C3F & 98.2 & \textbf{82.9} & \textbf{61.3} & 0.8 & \textbf{5.5} \\
         \midrule 
         \textbf{Standard (1094528)} & ViT & \textbf{98.6} & 63.4 & 0.0 & $1.4 \times 10^{13}$ & 9.3 \\
        %  GloRo & ViT &  &  & & & \\
         \textbf{CertViT (Ours)} & ViT & 97.8 & \textbf{64.2} & \textbf{54.5} & \textbf{1.2} & \textbf{8.4} \\
         \midrule
         \multicolumn{7}{c}{\textbf{CIFAR-10} ($\epsilon = 36/255$)} \\
         \midrule
         \textbf{Standard (2528096)} & 4C3F & \textbf{84.6} & 51.1 & 0.0 & $2.7\times 10^4$ & 8.2 \\
         \textbf{BCP} & 4C3F & 64.4 & 59.4 & 50.0 & 5.7 & 16.3 \\
         \textbf{GloRo} & 4C3F & 73.2 & 66.3 & 49.0 & 6.3 & 49.1 \\
         \textbf{Local-Lip} & 4C3F & 75.7 & 68.3 & 67.6 & 2.5 & 20.4 \\
         \textbf{CertViT (Ours)} & 4C3F & 81.2 & \textbf{69.8} & \textbf{69.1} & \textbf{1.9} & \textbf{7.4} \\
         \midrule 
         \textbf{Standard (2360672)} & 6C2F & \textbf{86.4} & 50.5 & 0.0 & $2.8\times10^{5}$ & 17.5 \\
         \textbf{BCP} & 6C2F & 65.7 & 60.8 & 51.3 & 11.35 & \textbf{35.0} \\
         \textbf{GloRo} & 6C2F & 70.7 & 63.8 & 49.3 & 9.21 & 140.1 \\
         \textbf{Local-Lip} & 6C2F & 69.8 & \textbf{64.3} & 54.1 & 7.89 & 43.7 \\
         \textbf{CertViT (Ours)} & 6C2F & 82.1 & 63.2 & \textbf{57.3} & \textbf{6.12} & 19.2 \\
         \midrule 
         \textbf{Standard (4086912)} & ViT & \textbf{81.4} & 33.7 & 0.0 & $8.2 \times 10^{16}$ & 331.5 \\
        %  GloRo & ViT &  &  & & & \\
        %  Local-Lip & ViT &  &  & & & \\
         \textbf{CertViT (Ours)} & ViT & 75.1 & \textbf{42.7} & \textbf{33.1} & \textbf{9.1} & \textbf{358.0} \\
         \midrule
         \multicolumn{7}{c}{\textbf{CIFAR-100} ($\epsilon = 36/255$)} \\
        %  \midrule
        %  Standard & 6C2F &  &  & & & \\
        %  GloRo & 6C2F &  &  & & & \\
        %  Local-Lip & 6C2F &  &  & & & \\
        %  CertViT (Ours) & 6C2F & & & &  & \\
         \midrule 
         \textbf{Standard (2436864)} & 8C2F&  \textbf{62.3} & 25.3 & 0.0 & $8.1\times 10^7$ & 128.6 \\
         \textbf{GloRo} & 8C2F & 29.3 & 27.7 & 21.3 & 10.2 & 514.3\\
         \textbf{Local-Lip} & 8C2F & 34.0 & 31.1 & 22.9 & 8.9 & 360.7 \\
         \textbf{CertViT (Ours)} & 8C2F & 42.4 & \textbf{35.2} & \textbf{25.4} & \textbf{7.2} & \textbf{77.1} \\
         \midrule 
         \textbf{Standard (4916736)} & ViT & \textbf{55.3} & 15.4 & 0.0 & $6.3 \times 10^{14}$ & 397.6 \\
        %  GloRo & ViT &  &  & & & \\
        %  Local-Lip & ViT &  &  & & & \\
         \textbf{CertViT (Ours)} & ViT & 46.2 & \textbf{21.6} & \textbf{9.1} & \textbf{12.1} & \textbf{429.4} \\
         \midrule
         \multicolumn{7}{c}{\textbf{TinyImageNet} ($\epsilon = 36/255$)} \\
         \midrule 
         \textbf{Standard (5257984)} & 8C2F & \textbf{39.1} & 12.1 & 0.0 & $2.9\times10^8$ & 583.0\\
         \textbf{GloRo} & 8C2F & 35.5 & 32.3 & 22.4 & 7.7 & 2331.8  \\
         \textbf{Local-Lip} & 8C2F & 37.4 & 34.2 & \textbf{27.4} & \textbf{5.9} & 1020.2 \\
         \textbf{CertViT (Ours)} & 8C2F & 38.1 & \textbf{34.9} & 26.3 & 6.1 & \textbf{349.8} \\
         \midrule 
         \textbf{Standard (9060864)} & ViT & \textbf{42.4} & 10.8 & 0.0 & $1.64 \times 10^{28}$ & 2860.1 \\
        %  GloRo & ViT &  &  & & & \\
        %  Local-Lip & ViT &  &  & & & \\
         \textbf{CertViT (Ours)} & ViT & 36.3 & \textbf{13.1} & \textbf{2.3} & \textbf{13.2} & \textbf{3088.9} \\
         \bottomrule
    \end{tabular}
    \caption{Comparison of CertViT with state-of-the-art Lipschitz bounding methods. We use Clean, PGD and Certified Accuracy as the performance metrics. We also report the Lipschitz constant values obtained and the FLOPs taken by each method. Note that for CertViT we do not consider the training FLOPs used to obtain the pre-trained network. In case of CIFAR-10 (6C2F) BCP takes less time (FLOPs) compared to standard + CertViT.}
    \label{tab:benchmarks}
    \vspace{-1em}
\end{table*}

\begin{table*}[!ht]
    \centering
    \scalebox{0.86}{
    \begin{tabular}{l c c c c c c c c c c c c c c}
    \toprule
    \multirow{2}{*}{\textbf{Model ($\mathbf{\rightarrow}$)}} & \multicolumn{7}{c}{\textbf{ViT}} & \multicolumn{3}{c}{\textbf{DeiT}} & \multicolumn{4}{c}{\textbf{Swin}} \\
   \cmidrule(lr){2-8}  \cmidrule(lr){9-11}  \cmidrule(lr){12-15}
         &  \textbf{T/16} & \textbf{S/16} & \textbf{S/32} & \textbf{B/8} & \textbf{B/16} & \textbf{B/32} & \textbf{L/16} & \textbf{T} & \textbf{S} & \textbf{B} & \textbf{T} & \textbf{S} & \textbf{B} & \textbf{L} \\
    \cmidrule(lr){2-8}  \cmidrule(lr){9-11}  \cmidrule(lr){12-15}
          \textbf{DP Params ($\mathbf{\times10^{6}}$)} & 5.7 & 22.0 & 22.9 & 86.4 & 80.5 & 88.2 & 304.1 & 5.7 & 22.0 & 86.4 & 28.3 & 49.5 & 87.7 & 196.4 \\
        %  DP FLOPs & 1.1 & 4.2 & 1.1 & 66.8 & 16.9 & 4.4 & 59.7 & 1.1 & 4.2 & 16.9 & 4.4 & 8.5 & 15.2 & 34.1 \\
         \textbf{L2 Params ($\mathbf{\times10^{6}}$)} & 5.2 & 20.2 & 21.0 & 78.8 & 79.2 & 81.0 & 278.4 & 5.2 & 20.2 & 78.8 & 25.7 & 44.9 & 79.9& 179.4 \\
         \textbf{L2 FLOPs ($\mathbf{\times 10^{16}}$)} & 6.4 &  25.0 &   6.4 & 392.7 & 100.6 & 
        25.6 & 350.4 &   6.4 &  26.7 & 396.8 & 26.3 &  50.1 &  90.3 & 199.9 \\
         \textbf{CertViT FLOPs ($\mathbf{\times 10^{16}}$)} &  7.7 &  31.2 &   7.7 & 471.2 & 120.7 &
        30.7 & 420.5 &   8.2 &  33.2 & 473.2 &
        31.5 &  59.9 & 108.4 & 239.8 \\
    \bottomrule
    \end{tabular}
    }
    \caption{Comparison of parameters of DP-attention and L2-attention transformers and the computation utilized to obtain the L2-attention from DP attention along with FLOPs used to constrain the L2-attention transformer using CertViT. \textbf{T/16} means \textbf{Tiny} variant with patch size $16$. \textbf{S}, \textbf{B}, and \textbf{L} are \textbf{Small}, \textbf{Base}, and \textbf{Large}, respectively.}
    \label{tab:vit_params}
    % \vspace{-1em}
\end{table*}

\begin{table*}[!ht]
      \centering
      \scalebox{0.9}{
      \begin{tabular}{l c c   c c c  c c c c }
      \toprule
      \multirow{2}{*}{\textbf{Model ($\downarrow$)}} &  \multicolumn{2}{c}{\textbf{DP Attention}} & \multicolumn{3}{c}{\textbf{L2 Attention}} &\multicolumn{4}{c}{\textbf{CertViT on L2 Attention}} \\
       \cmidrule{2-10}
       &  \textbf{Clean(\%)} & \textbf{PGD(\%)} &  \textbf{Clean(\%)} & \textbf{PGD(\%)} & \textbf{Lip. ($\mathbf{\times 10^{28}}$)} & \textbf{Clean(\%)} & \textbf{PGD(\%)} & \textbf{Cert.(\%)} & \textbf{Lip.} \\
       \cmidrule(lr){2-3}  \cmidrule(lr){4-6}  \cmidrule(lr){7-10}
       \textbf{ViT-T/16}   & 63.0 & 29.1 & 62.5 & 28.2 & 0.2 & 57.9 & 32.4  & 21.7 & 10.9 \\
       \textbf{ViT-S/16} & 74.2 & 48.2 & 73.7 & 47.3 & 1.1 & 69.2  & 51.3 & 0 & 72.9 \\
       \textbf{ViT-S/32}  & 67.6 & 40.6 & 67.5 & 39.5 & 2.3 & 63.1  & 44.5 & 0 & 317.4\\
       \textbf{ViT-B/8}  & 80.9 & 59.2 & 79.9 & 58.2 & 13.2 & 75.5  & 60.1 & 0 & 785.1 \\
       \textbf{ViT-B/16}  & 78.8 & 57.0 & 77.8 & 56.1 & 13.7 & 73.2   & 59.3 & 0 &  1720.4 \\
       \textbf{ViT-B/32} & 75.0 & 53.7 & 74.1 & 52.0 & 25.8 & 70.4  & 56.4 & 0 & 8791.7 \\
       \textbf{ViT-L/16} &  82.6 & 71.8 & 81.9 & 70.2 & $1.6 \times 10^{10}$ & 77.8  & 74.8 & 0 & $1.7 \times 10^{9}$\\
       \midrule
       \textbf{DeiT-T}   & 72.6 & 30.4 & 71.8 & 29.2 & 0.5 & 67.2   & 31.6 &  17.3 &  13.6 \\
       \textbf{DeiT-S}  & 71.5 & 45.4 & 70.5 & 44.8 & 1.7 & 64.6   & 49.3 & 0 &  93.7 \\
       \textbf{DeiT-B}  & 75.2 & 44.6 & 74.2 & 43.4 & 16.1 & 69.6  & 48.3 & 0 & 4592.0 \\
       \midrule
       \textbf{Swin-T }  & 72.6 & 21.2 & 71.2 & 20.5 & 7.2 & 65.8 & 22.8 & 0 &  483.6 \\
       \textbf{Swin-S }  & 75.7 & 26.0 & 74.3 & 25.1 & $3.8 \times 10^{9}$ & 69.7   & 28.0 & 0 & $2.7 \times 10^8$\\
       \textbf{Swin-B } & 79.8 & 30.1 & 78.5 & 29.1 & $11.7 \times 10^{10}$ & 73.6  & 31.1 & 0 & $1.9 \times 10^{10}$\\
       \textbf{Swin-L }  & 81.5 & 28.1 & 79.8 & 27.7 & $23.1 \times 10^{10}$ & 73.9  & 30.2 & 0 & $4.1 \times 10^{10}$\\
       \bottomrule
        \end{tabular}}
        \caption{Comparisons of different ViT variants pre-trained on ImageNet-1K adapted to have L2 attention and constrained using CertViT. Lip. denotes Lipschitz constant of the network.}
        \label{tab:vit_results}
\vspace{-1em}
\end{table*}

\section{Experiment} \label{sec:experiments}

We perform various experiments to demonstrate our proposed method's effectiveness in making neural network models more robust and provide a good trade-off between clean and certified accuracy. In order to manage our experiments we use Polyaxon\footnote{\url{https://github.com/polyaxon/polyaxon}} on a Kubernetes\footnote{\url{https://kubernetes.io/}} cluster and use 2 computing nodes with 16 A100 GPUs in each node (40GB VRAM per GPU). Experimental details are available in Appendix~\ref{appen:hyper-param}.

\subsection{Comparison to Existing Lipschitz Bounding Methods}

 To concertize the effectiveness of the proposed method in reducing the Lipschitz bounds, we first compare it with existing Lipschitz based methods used in certifying mid-size networks with convolution and dense layers. We compare our results with state-of-the-art of methods such as GloRo \cite{leino2021globally}, Local-Lip \cite{huang2021training}, and BCP \cite{lee2020lipschitz}. We show these results on limited-sized datasets: MNIST, CIFAR-10, CIFAR-100 and TinyImageNet in Table~\ref{tab:benchmarks}. We used the custom networks used in the certified robustness literature. i.e., 4C3F: MNIST, 4C3F and 6C2F for CIFAR-10, 8C2F for CIFAR-100 and TinyImageNet. We trained our model with the same configurations as provided in Local-Lip \cite{huang2021training} for fair comparisons. Next, we perform experiments on ViTs on each of the datasets. We train a 6-layer ViT on MNIST and CIFAR-10, 10-layer ViT on CIFAR-100, and 12-layer ViT on TinyImageNet. We replace DP attention with L2 attention to make architecture Lipschitz continuous, as explained in the previous sections. For all experiments of CertViT, we use $\beta=0.01$, $\eta=0.1$, and $\lambda=1.2$. More details about ViT patch size, number of heads, MLP ratio, and embedding dimension is available in the Appendix~\ref{appen:hyper-param}.

%Furthermore, we apply our proposed Lipschitz constraining method (CertViT) on these two fine-tuned networks. Similarly, we take two smaller variants for ViT (tiny and base), DeiT (tiny and small), and Swin (tiny and small) transformers. We replace DP with L2 attention in the ImageNet-1K pre-trained weights of the transformers as well and then apply our Lipschitz constraining method (CertViT) on them.

For each experiment, we report accuracy on non-perturbed inputs (clean accuracy), accuracy on adversarial perturbations generated via PGD attack \cite{madry2017towards} (PGD accuracy), and the proportion of inputs that can be correctly classified and certified within $\epsilon$-ball (certified accuracy). Certified accuracy gives a lower bound on the number of correctly classified points that are robust, and PGD accuracy serves as an upper bound on the same quantity. We also report the Lipschitz bounds for the trained models. We test our Lipschitz constrained neural networks to certify robustness against $\ell_2$ perturbations within an $\epsilon$-neighborhood of $1.58$ for MNIST, $36/255$ for CIFAR-10, CIFAR-100, and TinyImageNet (these are the $\ell_2$ norm bounds that have been commonly used in the previous literature.). We also tabulate the computational budget required to train each of these models in terms of FLOPs. More details on the training setup can be found in Appendix~\ref{appen:flow}. For different state-of-the-art, we used hyper-parameters mentioned in the respective works detailed in Appendix~\ref{appen:hyper-param} for details.

\textbf{Observations}: As mentioned earlier, unconstrained neural network models have very high Lipschitz bounds. We observe that ViTs with just 1M parameters have very high Lipschitz constant of order $10^{13}$, making them unsuitable for certification. The high value can be attributed to the multi-headed self-attention layer in the transformers and residual branches. Our proposed method, CertViT is very successful in lowering the Lipschitz bounds of networks compared to the existing works on networks with convolution and dense layers. CertViT also maintained accuracy close to the pre-trained network. We achieved better PGD and certified accuracy than existing works for all the datasets. CertViT also takes less computational time than other methods, which can be attributed to its applicability to pre-trained weights. Moving to ViTs, CertViT successfully lowered the Lipschitz bounds of pre-trained classifiers by orders of magnitude in all the cases, and we were able to certify the samples. It is can also be observed that the performance of existing methods on ViTs is unknown, and the theory usually revolves around non-expansive operators such as ReLU. We still tried these methods on ViTs using a few modifications and found they failed miserably in lowering Lipschitz bounds and hence did not certify any samples. Therefore, we have removed them from our comparisons. In contrast, our method was able to tighten the Lipschitz bound and improve PGD and certified accuracy. In the case of transformers, too CertViT requires considerably fewer FLOPs than the existing works implying it requires fewer epochs to converge.

\subsection{Constraining Lipschitz of large pre-trained transformers}

To show the applicability of our proposed method on large transformers, we apply it on ImageNet-1K pre-trained weights of ViT (tiny, small, base, and large), DeiT (tiny, small, and base), and Swin (tiny, small, base, and large). All weights were obtained from timm\footnote{\url{https://github.com/rwightman/pytorch-image-models}} library. First, we replace the DP attention in ViT and DeiT networks with L2 attention. Similarly, in Swin, we replace the shifted window and window attention with their L2 variants as described in Appendix~\ref{appen:swin}. We then apply our Lipschitz constraining method (CertViT) on these L2-adapted networks. Table~\ref{tab:vit_params} shows the number of parameters for DP attention and L2 attention versions of all the large transformers used in our experiments. The Table also reports the computational cycles (FLOPs) used in adapting pre-trained  DP attention to L2 attention and the computational cycles used by CertViT when constraining L2 attention pre-trained networks. Table~\ref{tab:vit_results} shows results obtained by DP attention, L2 attention, and Lipschitz constrained (CertViT) L2 attention networks. Our proposed method managed to constrain the Lipschitz constant while maintaining acceptable clean and PGD. We were only able to obtain certified accuracy for ViT-T/16 and DeiT-T. This is because we are calculating a very loose upper Lipschitz bound that increases with the depth and parameters of the network. In all other cases, we reduced Lipscthiz, leading to increased PGD accuracy. For PGD accuracy on ImageNet-1K, we randomly select 1000 samples such that each class has one sample from the test set and report the accuracy of the selected sample. The value of $\epsilon$-neighbourhood is kept at $36/255$.

\section{Conclusion} \label{sec:conclusion}
In this work, we propose an efficient way of providing certified robustness using Lipschitz bounds focusing on vision transformers. With the ever-increasing deployment of neural networks, such formal guarantees are necessary for safety-critical applications. Such Lipschitz constrained training has been missing in the literature for transformers. We have presented a proximal projection step for lowering the Lipschitz bounds while maintaining the accuracy of the classifier. We show the efficacy of CertViT on simple architectures with convolution and fully-connected layers compared to existing techniques. We establish this pipeline for making the transformer layers Lipschitz continuous by using $L2$ attention layers. In the case of pre-trained transformers, we have adapted DP attention to L2 attention layers. We use our proposed method, CertViT, for pre-trained transformers. Results obtained on large transformer variants pre-trained on ImageNet-1K show the efficacy of our proposed method. 
%We use a relaxed Lipschitz constrained training that can be used to train robust models, such training strategy won't over-regularize the model and hence won't drop the clean accuracy drastically. For pre-trained networks, we formulate an optimization problem that alternates between lowering the Lipschitz constant and maintaining the clean accuracy. We show the results on widely used variants of ViTs and datasets. 

%\section{Broader Impact}

\newpage
\bibliographystyle{ieeetr}
\bibliography{refs}

\clearpage
\newpage
\appendix
\section{Appendix}
\addcontentsline{toc}{section}{Appendices}
\renewcommand{\thesubsection}{\Alph{subsection}}
\subsection{GELU activation} \label{gelu}
The Lipschitz constant of any function $f(x)\colon \RR \to \RR$, which follows Lipschitz continuity, can be calculated as the absolute maximum value of its derivative $ f'(x)$. That is $|f'(x)| \leq L$ $\forall x$ then $L$ is the Lipschitz constant of $f$. GELU activation function ($f(x)$) and its derivative ($f'(x)$)  is given as
\begin{align}
  f(x) & = \frac{x}{2} [ 1 + erf(x/\sqrt{2})] \\
& \text{where,} \quad erf(x) = \frac{2}{\sqrt{\pi}}\int_{0}^{x} e^{-t^2 dt}
\end{align}
\vspace{1em}
\begin{align}
    f'(x) = \frac{xe^{-x^2/2}}{\sqrt{2\pi}} + \frac{erf(x/\sqrt{2})}{2} + \frac{1}{2}
\end{align}

\begin{figure}[!h]
    \centering
    \includegraphics[scale=0.59]{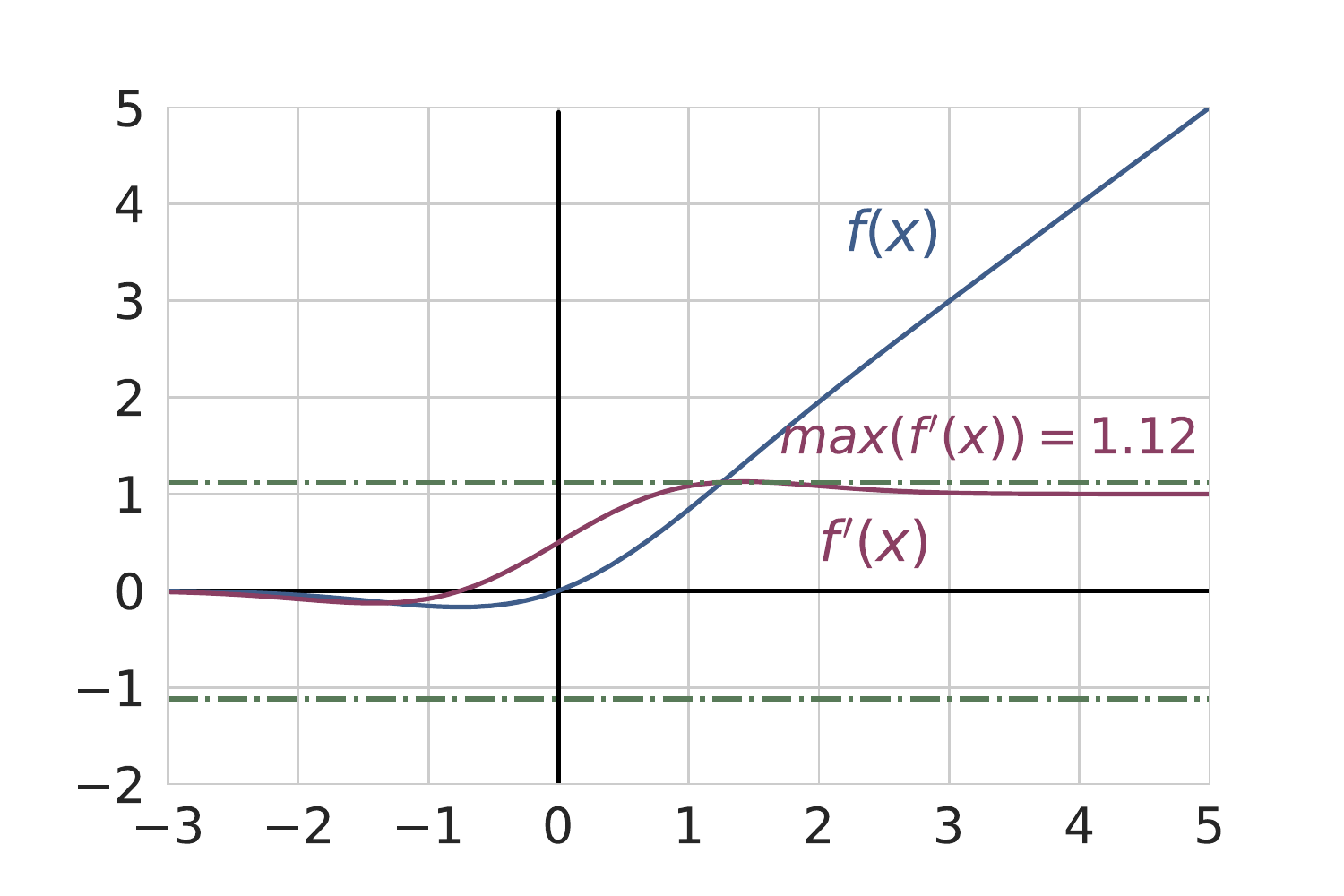}
    \caption{ Representation of GELU activation function and its derivative.  The maximum absolute value of the derivative is $1.12$, as seen from the graph. }
    \label{fig:gelu}
\end{figure}

From the figure ~\ref{fig:gelu} we see that the maximum of $|f'(x)|$ is 1.12. So, for the GELU activation function in the transformers, we loosely take the Lipschitz constant as 1.12 and the Lipschitz constant of the MLP layer with the GELU activation function as $1.12 \|W\|_s$.

\subsection{Training Setup}\label{appen:flow}

\subsubsection{Training Strategies}
\begin{figure}[h]
    \centering
    \includegraphics[scale=0.5]{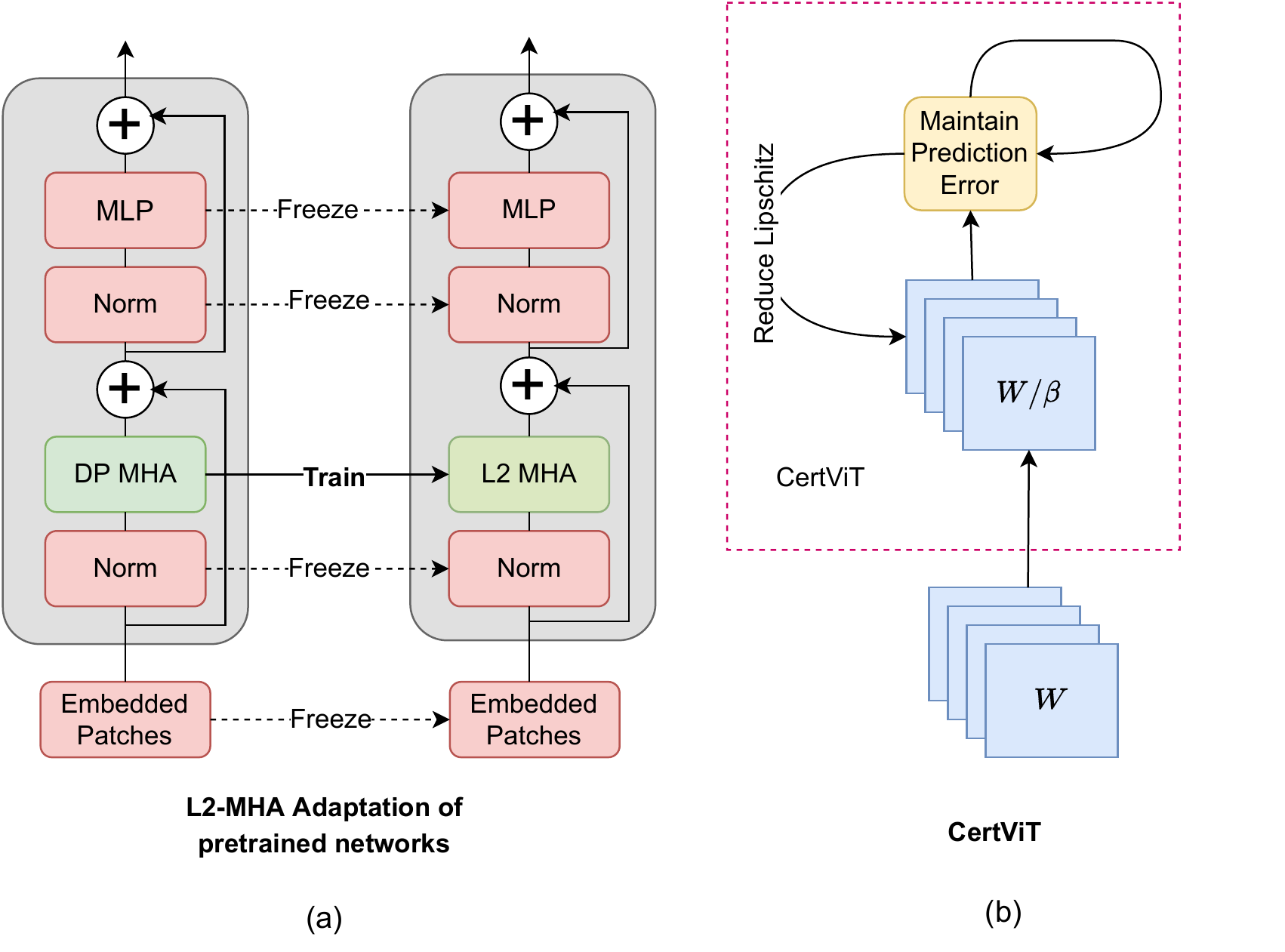}
    % \caption{(a) Training from scratch, constrained Lipschitz training, the weights of the network are divided by its spectral normalization ($\sigma$). This is further relaxed by the parameter $\gamma$ (b) L2 adaptation in a pre-trained network that was trained with DP-MHA. All the layers in the transformer encoder (shown in red) are frozen and the DP-MHA is replaced with L2 MHA and trained for adaptation. (c) Depiction of the proposed approach CertViT which alternates between maintaining a good accuracy and a small Lipschitz constant.}
    \caption{(a) L2 adaptation in a pre-trained network that was trained with DP-MHA. All the layers in the transformer encoder (shown in red) are frozen and the DP-MHA is replaced with L2 MHA and trained for adaptation. (b) Depiction of the proposed approach CertViT, which alternates between maintaining a good accuracy and a small Lipschitz constant.}
    \label{fig:training_pipe}
\end{figure}

Figure~\ref{fig:training_pipe} describes the different training strategies and pipelines we have used in our experiments. 

\subsubsection{Experimental Setup}
PyTorch is employed to implement all the experiments. We use \textit{timm}\footnote{\url{https://github.com/rwightman/pytorch-image-models}} implementation of ViT\footnote{\url{https://github.com/rwightman/pytorch-image-models/blob/0.5.x/timm/models/vision_transformer.py}} and Swin\footnote{\url{https://github.com/rwightman/pytorch-image-models/blob/0.5.x/timm/models/swin_transformer.py}}. We replace Linear and Convolutional layers with our own implementation called LinearX and ConvX, respectively. These layers allow us to compute the Lipschitz constant and project weight according to the requirement of CertViT. We also implement L2 attention to replace attention in ViT and Swin. 
% Code snippets for LinearX, ConvX, L2Attention, and single iteration of CertViT are listed in Appendix~\ref{appen:codes}. 
In order to manage our experiments we use Polyaxon\footnote{\url{https://github.com/polyaxon/polyaxon}} on a Kubernetes\footnote{\url{https://kubernetes.io/}} cluster and use 2 computing nodes with 16 A100 GPUs in each node (40GB VRAM per GPU).

\subsection{Hyper-parameter} \label{appen:hyper-param}

\subsubsection{Architectures}

\begin{table}[!h]
      \centering
      \scalebox{0.75}{
      \begin{tabular}{l c c c c c c}
      \toprule
      \multirow{2}{*}{\textbf{Dataset}} & \textbf{Input} &  \multirow{2}{*}{\textbf{Channels}} & \textbf{Patch} & \multirow{2}{*}{\textbf{Layers}} & \textbf{Embed}  & \textbf{MLP}  \\
      & \textbf{Size} & & \textbf{Size} & & \textbf{Dim} & \textbf{Ratio} \\
      \midrule
      \textbf{MNIST} & 28 & 1 & 7 & 6 & 128 & 4 \\
      \textbf{CIFAR-10} & 32 & 3 & 4 & 6 & 192 & 3 \\
      \textbf{CIFAR-100} & 32 & 3 & 4 & 12 & 192 & 3 \\
      \textbf{TinyImageNet} & 64 & 3 & 4 & 12 & 384 & 3 \\
      \bottomrule
        \end{tabular}}
        \caption{ViT architectures used for different datasets when trained from scratch.}
        \label{tab:toy_vit}
\end{table}

\begin{table}[!h]
      \centering
      \scalebox{0.75}{
      \begin{tabular}{ c c c c c c c}
      \toprule
      \multirow{2}{*}{\textbf{Name}} & \textbf{Input} &  \multirow{2}{*}{\textbf{Channels}} & \textbf{Patch} & \multirow{2}{*}{\textbf{Layers}} & \textbf{Embed} & \textbf{MLP} \\
      & \textbf{Size} & & \textbf{Size} & & \textbf{Dim} & \textbf{Ratio} \\
      \midrule
      \textbf{ViT-T/16} & 224 & 3 & 16 & 12 & 192 & 3 \\
      \textbf{ViT-S/16} & 224 & 3 & 16 & 12 & 384 & 6 \\
      \textbf{ViT-S/32} & 224 & 3 & 32 & 12 & 384 & 6 \\
      \textbf{ViT-B/8} & 224 & 3 & 8 & 12 & 768 & 12 \\
     \textbf{ViT-B/16} & 224 & 3 & 16 &  12 & 768 & 12\\
      \textbf{ViT-B/32} & 224 & 3 & 32 & 12 & 768 & 12 \\
      \textbf{ViT-L/16} & 224 & 3 & 16 & 24 & 1024 & 16 \\
      \midrule
      \textbf{DeiT-T} & 224 & 3 & 16 & 12 & 192 & 3 \\ 
      \textbf{DeiT-S} & 224 & 3 & 16 & 12 & 384 & 6 \\
      \textbf{DeiT-B} & 224 & 3 & 16 &  12 & 768 & 12  \\
      \bottomrule
        \end{tabular}}
        \caption{ViT architectures are used for different datasets when ImageNet-1K pre-trained weights are used.}
        \label{tab:vit_architectures_pretrained}
\end{table}

\begin{table}[!h]
      \centering
      \scalebox{0.63}{
      \begin{tabular}{c c c c c c c c}
      \toprule
      \multirow{2}{*}{\textbf{Name}} & \textbf{Input} &  \multirow{2}{*}{\textbf{Chnls}} & \textbf{Patch} & \textbf{Window} &  \multirow{2}{*}{\textbf{Layers}} & \textbf{Embed} & \multirow{2}{*}{\textbf{Heads}} \\
      & \textbf{Size} & & \textbf{Size} & \textbf{Size} & & \textbf{Dim} &  \\
      \midrule
      \textbf{Swin-T} & 224 & 3 & 4 & 7 & 2, 2, 6, 2 & 96 & 3, 6, 12, 24\\
      \textbf{Swin-S} & 224 & 3 & 4 & 7 & 2, 2, 18, 2 & 96 & 3, 6, 12, 24 \\
      \textbf{Swin-B} & 224 & 3 & 4 & 7 & 2, 2, 18, 2 & 128 & 4, 8, 16, 32 \\
      \textbf{Swin-L} & 224 & 3 & 4 & 7 & 2, 2, 18, 2 & 192 & 6, 12, 24, 48 \\
      \bottomrule
        \end{tabular}}
        \caption{Swin architectures used for ImageNet-1K.}
        \label{tab:swin_architectures}
\end{table}

Table~\ref{tab:toy_vit} shows ViT architecture parameters used to train from scratch on MNIST, CIFAR-10, CIFAR-100 and TinyImageNet datasets. Table~\ref{tab:vit_architectures_pretrained} shows ViT and DeiT variants parameters for ImageNet-1K pre-trained networks used in the experiment. Swin variants (pre-trained on ImageNet-1k) details is shown in Table~\ref{tab:swin_architectures}.

\subsubsection{Training from scratch}

\begin{table}[!h]
      \centering
      \scalebox{0.75}{
      \begin{tabular}{l c c c c}
      \toprule
      \textbf{Dataset} & \textbf{MNIST} & \textbf{CIFAR-10} & \textbf{CIFAR-100} & \textbf{TinyImageNet} \\
      \midrule
      \textbf{Warm-up}  & 5 & 15 & 30 & 35 \\
      \textbf{Batch Size} & 512 & 512 & 512 & 512 \\  
      \textbf{Epochs} & 500 & 800 & 800 & 500 \\ 
      \textbf{$\epsilon_{\text{train}}$} & 1.74 & 0.16 & 0.16 & 0.16\\
      \textbf{$\epsilon_{\text{test}}$} & 1.58 & 0.141 & 0.141 & 0.141 \\
      \textbf{Optimizer} & Adam & Adam & Adam & Adam \\
      \textbf{Init LR} & 1e-3 & 1e-3 & 1-e3 & 2.5e-4 \\
      \textbf{LR Decay} & 5e-6 & 5e-6 & 5e-6 & 5e-7\\
      \textbf{$\epsilon_{\text{schedule}}$} & single & single & single & single\\
      \textbf{Power Iter} & 5 & 5 & 5 & 5 \\
      \bottomrule
        \end{tabular}}
        \caption{Hyperparameters used for training GloRo.}
        \label{tab:gloro_hyperparams}
\end{table}

% \begin{table*}[!h]
%       \centering
%       \scalebox{0.75}{
%       \begin{tabular}{l c c c c c c c c c c}
%       \toprule
%       Dataset & Warm-up & Batch Size & Epochs & $\epsilon_{\text{train}}$ & $\epsilon_{\text{test}}$ & Optimizer & Init LR & LR Decay & $\epsilon_{\text{schedule}}$ & Power Iter\\
%       \midrule
%       MNIST & 5 & 512 & 500 & 1.74 & 1.58  & adam & 1e-3 & 5e-6 & single & 5 \\ 
%       CIFAR-10 & 15 & 512 & 800 & 0.16 & 0.141  & adam & 1e-3 & 5e-6 & single & 5 \\ 
%       CIFAR-100 & 30 & 512 & 800 & 0.16 & 0.141  & adam & 1e-3 & 5e-6 & single & 5 \\
%       TinyImageNet & 35 & 512 & 1000 & 0.16 & 0.141 & adam & 1e-3 & 5e-6 & single & 5 \\
%       \bottomrule
%         \end{tabular}}
%         \caption{Hyperparameters used for training GloRo.}
%         \label{tab:gloro_hyperparams}
% \end{table*}

\begin{table}[!h]
      \centering
      \scalebox{0.72}{
      \begin{tabular}{l c c c c}
      \toprule
      \textbf{Dataset} & \textbf{MNIST} & \textbf{CIFAR-10} & \textbf{CIFAR-100} & \textbf{TinyImageNet}\\
      \midrule
      \textbf{Warm-up}  & 0 & 5 & 20 & 30 \\
      \textbf{Batch Size} & 256 & 256 & 256 & 128 \\  
      \textbf{Epochs} & 300 & 300 & 800 & 250 \\ 
      \textbf{$\epsilon_{\text{train}}$} & 1.58 & 0.1551 & 0.1551 & 0.16\\
      \textbf{$\epsilon_{\text{test}}$} & 1.58 & 0.141 & 0.141 & 0.141 \\
      \textbf{Init LR} & 1e-3 & 1e-3 & 1-e3 & 2.5e-4 \\
      \textbf{End LR} & 5e-6 & 5e-6 & 5e-6 & 5e-7\\
      \textbf{$\lambda_{sparse}$} & 0.0 & 0.0 & 0.0 & 0.01 \\
      \textbf{$\lambda_{\theta}$} & 0.0 & 0.0 & 0.0 & 0.1 \\
      \textbf{LR Decay Epoch} & 150 & 200 & 400 & 150 \\
      \textbf{$\epsilon_{\text{schedule}}$ Epochs} & 150 & 200 & 400 & 125 \\
      \textbf{Power Iter} & 5 & 5 & 2 & 1 \\
      \bottomrule
        \end{tabular}}
        \caption{Hyperparameters used for training Local-Lip}
        \label{tab:locallip_hyperparams}
\end{table}

% \begin{table*}[!h]
%       \centering
%       \scalebox{0.75}{
%       \begin{tabular}{l c c c c c c c c c c c c}
%       \toprule
%       Dataset & Warm-up & Batch Size & Epochs & $\epsilon_{\text{train}}$ & $\epsilon_{\text{test}}$  & Init LR & End LR & $\lambda_{sparse}$ & $\lambda_{\theta}$ & LR Decay Epoch & $\epsilon_{\text{schedule}}$ Epochs & Power Iter\\
%       \midrule
%       MNIST & 0 & 256 & 300 & 1.58 & 1.58  & 1e-3 & 5e-6 & 0.0 & 0.0 & 150 & 150 & 5 \\ 
%       CIFAR-10 & 5 & 256 & 300 & 0.1551 & 0.141  & 1e-3 & 5e-6 & 0.0 & 0.0 & 200 & 200 & 5 \\
%       CIFAR-100 & 20 & 256 & 800 & 0.1551 & 0.141  & 1e-3 & 5e-6 & 0.0 & 0.0 & 400 & 400 & 2 \\
%       TinyImageNet & 30 & 128 & 250 & 0.16 & 0.141 & 2.5e-4 & 5e-7 & 0.01 & 0.1 & 150 & 125 & 1 \\
%       \bottomrule
%         \end{tabular}}
%         \caption{Hyperparameters used for training Local-Lip}
%         \label{tab:locallip_hyperparams}
% \end{table*}

\begin{table}[!h]
      \centering
      \scalebox{0.7}{
      \begin{tabular}{l c c c c}
      \toprule
      \textbf{Dataset} & \textbf{MNIST} &\textbf{CIFAR-10} & \textbf{CIFAR-100} & \textbf{TinyImage}\\
      \midrule
      \textbf{Batch Size}  & 512 & 512 & 512 & 512 \\ 
      \textbf{Epochs} &  100 & 200 & 200 & 200 \\ 
      \textbf{Optimizer} & Adam & Adam & Adam & Adam \\ 
      \textbf{Init LR} & 1e-3 & 1e-3 & 1e-3 & 2.5e-4\\
      \textbf{LR Decay} & 5e-6 & 5e-6 & 5e-6 & 5e-7\\
      \textbf{Scheduler} & MultiStep & CosAneal & CosAneal & CosAnneal\\
      \textbf{Milestones} & (50, 60, 70, 80)  & NA & NA & NA \\
      \textbf{$\gamma_{\text{scheduler}}$} & 0.2 & NA & NA & NA\\
      \textbf{$\eta_{\text{min}}$} & NA & 1e-5 & 1e-5 & 1e-5 \\
      \textbf{CertViT Epochs} & 5 & 5 & 5 & 5 \\
      \textbf{$\mathbf{\proj_C}$ Epochs} & 2 & 2 & 2 & 2 \\
      \textbf{$\eta$} & 1e-2 & 1e-2 & 1e-2 & 1e-2 \\
      \textbf{$\beta$} & 0.1 & 0.1 & 0.1 & 0.2 \\
      \textbf{$\lambda$} & 1.1 & 1.2 & 1.2 & 1.2 \\ 
      \bottomrule
        \end{tabular}}
        \caption{Hyperparameters used for training transformers with L2 attention and then using CertViT to constrain the trained network.}
        \label{tab:l2_hyperparams}
\end{table}

Table~\ref{tab:gloro_hyperparams} and Table~\ref{tab:locallip_hyperparams}.show hyperparameters used in GloRo and LocalLip training of convolutional networks. We use TensorFlow implementation provided here\footnote{\url{https://github.com/klasleino/gloro},\url{https://github.com/yjhuangcd/local-lipschitz}}. 
% Table~\ref{tab:l2_gamma_hyperparams} shows hyper-parameters for constrained training $\mathcal{L}_{\gamma}$ of a network from scratch. Different warm-up and relaxation parameter ($\gamma$) explored is shown in the last two columns, respectively. 
Table~\ref{tab:l2_hyperparams} shows the hyperparameters required to train convolutional and ViTs on MNIST, CIFAR-10/100, and TinyImageNet from scratch.
%Detailed comparisons are available in Figure~\ref{fig:relax}.

\subsubsection{Constraining pre-trained networks}

% \begin{table*}[!h]
%       \centering
%       \scalebox{0.75}{
%       \begin{tabular}{l c c c c c c c}
%       \toprule
%       Dataset & Batch Size & Fine-tuning Epochs & Adaptation Epochs & CertViT Epochs & $\proj_C$ Epochs & $\eta$ & $\beta$  \\
%       \midrule
%       CIFAR-10 & 512 & 200 & 20 & 5 & 10 & 1e-7 & 1.2  \\ 
%       CIFAR-100 & 512 & 100 & 50 & 7 & 20 & 1e-7 & 1.2  \\ 
%       \bottomrule
%         \end{tabular}}
%         \caption{Hyperparameters used for fine-tuning ViT-T/16 on CIFAR-10 and CIFAR-100 followed by L2 adaptation and then constrained using CertViT.}
%         \label{tab:pretrianing_hyperparams}
% \end{table*}

% Hyper-parameters used in fine-tuning, L2 adaptation, and constraining using CertViT of ViT-T/16 (pre-trained on ImageNet-1K) for CIFAR-10 and CIFAR-100 datasets is shown in Table~\ref{tab:pretrianing_hyperparams}. The same batch size is used across all three tasks. $\eta$ is the error rate tolerance in the output of constrained layers. $\beta$ is the relaxation parameter used in CertViT.

\begin{table}[!h]
      \centering
      \scalebox{0.72}{
      \begin{tabular}{l c c c c c c c}
      \toprule
      \multirow{2}{*}{\textbf{Dataset}} & \textbf{Batch}  & \textbf{Adaptation} & \textbf{CertViT} & $\mathbf{\proj_C}$ & \multirow{2}{*}{\textbf{$\eta$}} & \multirow{2}{*}{\textbf{$\beta$}}  & \multirow{2}{*}{\textbf{$\lambda$}}\\
      & \textbf{Size} & \textbf{Epochs} & \textbf{Epochs} & \textbf{Epochs} &  & &  \\
      \midrule
      \textbf{ViT-T/16} & 2048 & 50 & 5 & 2 & 1e-2 & 0.1 & 1.2\\ 
      \textbf{ViT-S/16} & 2048 & 50 & 5 & 2 & 1e-2 & 0.1 & 1.2 \\
      \textbf{ViT-S/32} & 1024 & 70 & 5 & 2 & 1e-2 & 0.1 & 1.2 \\
      \textbf{ViT-B/8} & 128 & 70 & 6 & 2 & 1e-2 & 0.2 & 1.3 \\
      \textbf{ViT-B/16} & 256 & 50 & 5 & 2 & 1e-2 & 0.2 & 1.2 \\
      \textbf{ViT-B/32} & 256 & 50 & 6 & 2 & 1e-2 & 0.2 & 1.3 \\
      \textbf{ViT-L/16} & 256 & 70 & 6 & 2 & 1e-2 & 0.2 & 1.3 \\
      \hline 
      \textbf{DeiT-T} & 2048 & 50 & 5 & 2 & 1e-2 & 0.1 & 1.2 \\
      \textbf{DeiT-S} & 2048 & 70 & 6 & 2 & 1e-2 & 0.2 & 1.2 \\
      \textbf{DeiT-B} & 256 & 70 & 6 & 2 & 1e-2 & 0.2 & 1.3 \\
      \hline 
      \textbf{Swin-T} & 1024 & 40 & 5 & 2 & 1e-2 & 0.1 & 1.2 \\
      \textbf{Swin-S} & 512 & 50 & 6 & 2 & 1e-2 & 0.2 & 1.2 \\
      \textbf{Swin-B} & 256 & 50 & 6 & 3 & 1e-2 & 0.25 & 1.3 \\
      \textbf{Swin-L} & 128 & 70 & 7 & 3 & 1e-2 & 0.25 & 1.3  \\
      \bottomrule
        \end{tabular}}
        \caption{Hyperparameters used for L2 adaptation and CertViT applied on different transformer architectures pre-trained on ImageNet-1K.}
        \label{tab:pretrianing_hyperparams}
\end{table}

Hyper-parameters used in L2 adaptation and constraining using CertViT of transformer variants pre-trained on ImageNet-1K are shown in Table~\ref{tab:pretrianing_hyperparams}. 

\subsection{Swin} \label{appen:swin}

The Swin transformer builds hierarchical feature maps by merging image patches in deeper layers and has linear computation complexity to input image size due to computation of self-attention only within each local window, unlike in other ViTs where the attention is calculated globally and hence the complexity is quadratic with respect to patch number. Global self-attention in most ViTs variants is generally unaffordable for high-resolution images, while window-based self-attention is scalable. Window-based self-attention lacks connections across windows, limiting its modeling power. They propose a shifted window partitioning that introduces a connection between neighboring non-overlapping windows in consecutive Swin transformer blocks. 

In simple terms, Swin transformers calculate self-attention in non-overlapping local windows and are suitable for applications where the image resolution is high. The windows are arranged to partition the image in a non-overlapping manner evenly. The Swin transformers block is built by replacing the standard multi-head self-attention (MSA) module in the standard transformer block with a module based on shifted windows (W-MSA), with other layers kept the same. This W-MSA layer is still calculated using standard Dot product attention. We replace the Dot product in each W-MSA layer with L2 attention to make the layer Lipschitz continuous and everything else in the architecture remains the same.

\end{document}